\documentclass[preprint,12pt,authoryear]{elsarticle}

\usepackage{amssymb}
\usepackage[bookmarks,bookmarksnumbered]{hyperref}
\hypersetup{
    colorlinks = true,
    linkcolor = blue,
    anchorcolor =red,
    citecolor = blue,
    filecolor = red,urlcolor = red,
    pdfauthor=author
}

\usepackage{booktabs}
\usepackage{multirow}

\usepackage{graphicx}
\usepackage{float}
\usepackage{subfig}

\usepackage{makecell}

\usepackage{bm}
\usepackage{amsmath}

\usepackage{algorithm}
\usepackage{algorithmic}
\usepackage{diagbox}
\usepackage{graphicx}
\usepackage{amsmath}
\usepackage{graphics}
\usepackage{amsfonts}
\usepackage{amssymb}%
\usepackage{amsthm}
\usepackage{epsfig}
\usepackage{subfloat}
\usepackage{color}
\usepackage{listings}
\usepackage{booktabs}
\usepackage{multirow}
\usepackage{amsfonts}
\usepackage{multicol}
\usepackage{epstopdf}

\usepackage{amssymb}

\begin{document}

\begin{frontmatter}



\title{STAR: Zero-Shot Chinese Character Recognition with \\Stroke- and Radical-Level Decompositions }


 \author[ind1]{Jinshan Zeng\corref{cor1}}
    \author[ind1]{Ruiying Xu\corref{cor2}}
    \author[ind1]{Yu Wu}
    \author[ind1]{Hongwei Li}
    \author[ind1]{Jiaxing Lu}

    \address[ind1]{School of Computer and Information Engineering, Jiangxi Normal University, Nanchang, China, 330022.}
    \cortext[cor1]{E-mail: jinshanzeng@jxnu.edu.cn}


\begin{abstract}
Zero-shot Chinese character recognition has attracted rising attention in recent years. Existing methods for this problem are mainly based on either certain low-level stroke-based decomposition or medium-level radical-based decomposition. Considering that the stroke- and radical-level decompositions can provide different levels of information, we propose an effective zero-shot Chinese character recognition method by combining them. The proposed method consists of a training stage and an inference stage. In the training stage, we adopt two similar encoder-decoder models to yield the estimates of stroke and radical encodings, which together with the true encodings are then used to formalize the associated stroke and radical losses for training.  A similarity loss is introduced to regularize stroke and radical encoders to yield features of the same characters with high correlation. In the inference stage, two key modules, i.e., the stroke screening module (SSM) and feature matching module (FMM) are introduced to tackle the deterministic and confusing cases respectively. In particular, we introduce an effective stroke rectification scheme in FMM to enlarge the candidate set of characters for final inference. Numerous experiments over three benchmark datasets covering the handwritten, printed artistic and street view scenarios are conducted to demonstrate the effectiveness of the proposed method. Numerical results show that the proposed method outperforms the state-of-the-art methods in both character and radical zero-shot settings, and maintains competitive performance in the traditional seen character setting.
\end{abstract}




\end{frontmatter}


\section{Introduction}
\label{Introduction}
\begin{figure*}[ht]
    \vspace{0cm}
    \setlength{\abovecaptionskip}{-10pt} \setlength{\belowcaptionskip}{-1pt}
    \begin{center}
        \includegraphics[width=0.9\textwidth]{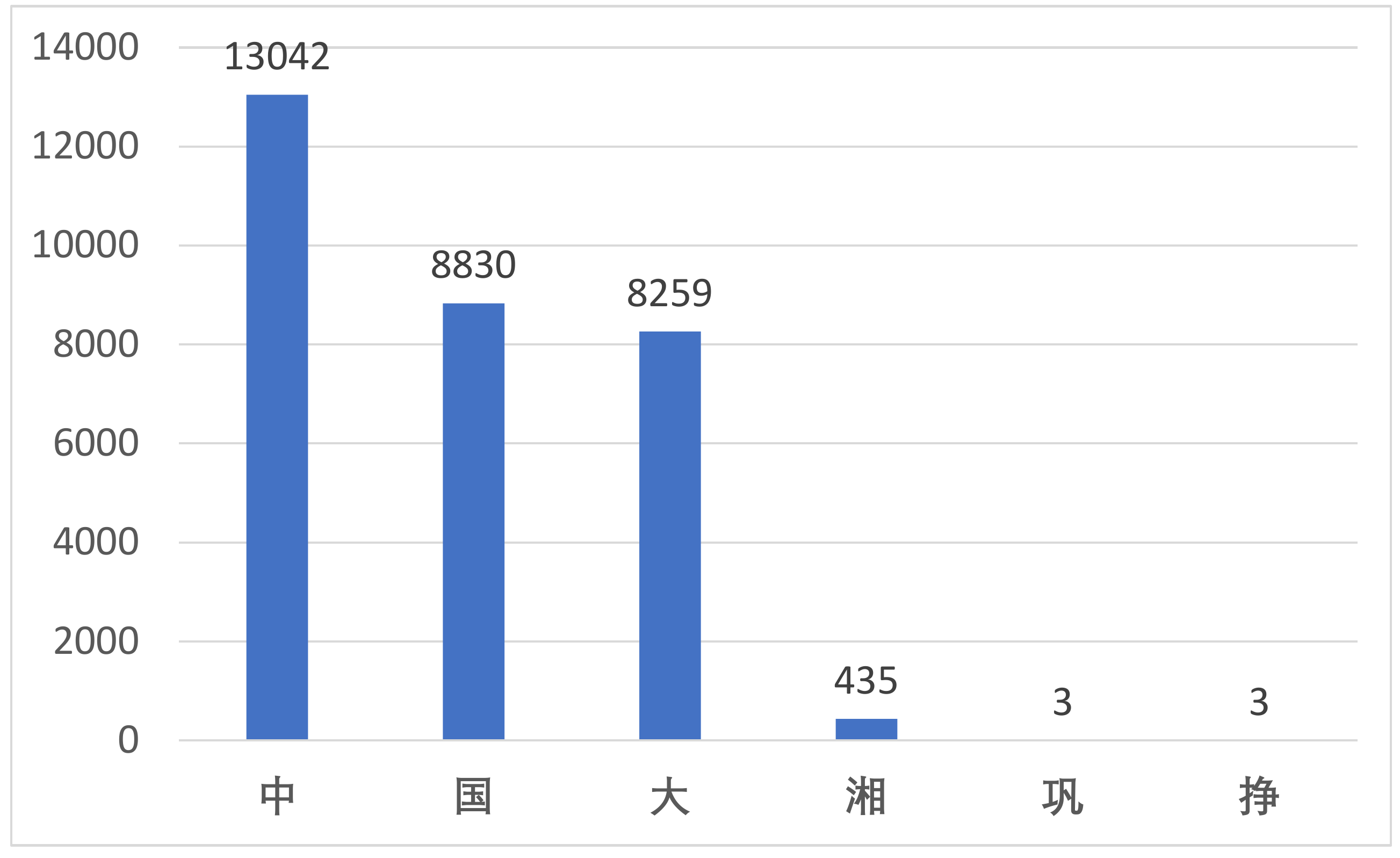}
    \end{center}
    \caption{The occurrence frequencies of six Chinese characters in the CTW dataset. It can observed that the occurrence frequencies of these Chinese characters are very different, and that some characters rarely appear.}
    \label{fig:Class_imbalance}
\end{figure*}

Chinese character recognition has been studied for decades \citep{feature_trier_1996,Normalization_liu_2003,gradient_feature_maps_chang_2006,end_wang_2011,MCDNN_2013,similar_tao_2014,DenseRAN_wang_2018}. Traditional approaches for Chinese character recognition can only recognize the categories of Chinese characters that appear in the training set and rely on a large amount of training data \citep{online_liu_2004,RAN_zhang_2018}. Existing benchmark Chinese character datasets such as the HWDB \citep{HWDB_liu_2013} and the CTW databases \citep{CTW_yuan_2019} collect roughly 3800 Chinese character categories, while there are a total of 70,244 Chinese characters according to the latest standard GB18030-2005. Thus, it is costly to collect and annotate a whole set of Chinese characters. Moreover, as shown in Figure \ref{fig:Class_imbalance}, the occurrence frequencies of characters in realistic scenarios are very different, 
and some characters even rarely appear, resulting in the limited performance of existing models for recognizing these characters with low occurrence frequencies \citep{SLD_chen_2021,REZCR_Diao_2022}. Therefore, the zero-shot Chinese character recognition, i.e., recognizing unseen characters by learning from sparse samples or training on a limited class of characters is an important and challenging task \citep{HDE_cao_2020}, and has been involved in many real-world applications such as autonomous driving \citep{driving_zhang_2021,drive_qian_2015}, historical documents retrieval \citep{document_maekawa_2019,obc306_huang_2019}, and handwritten signature identification \citep{signature_poddar_2020,signature_st_2020}.

\begin{figure*}[ht]
\centering
 \begin{minipage}{0.49\linewidth}
  \centering
  \includegraphics[width=.9\linewidth]{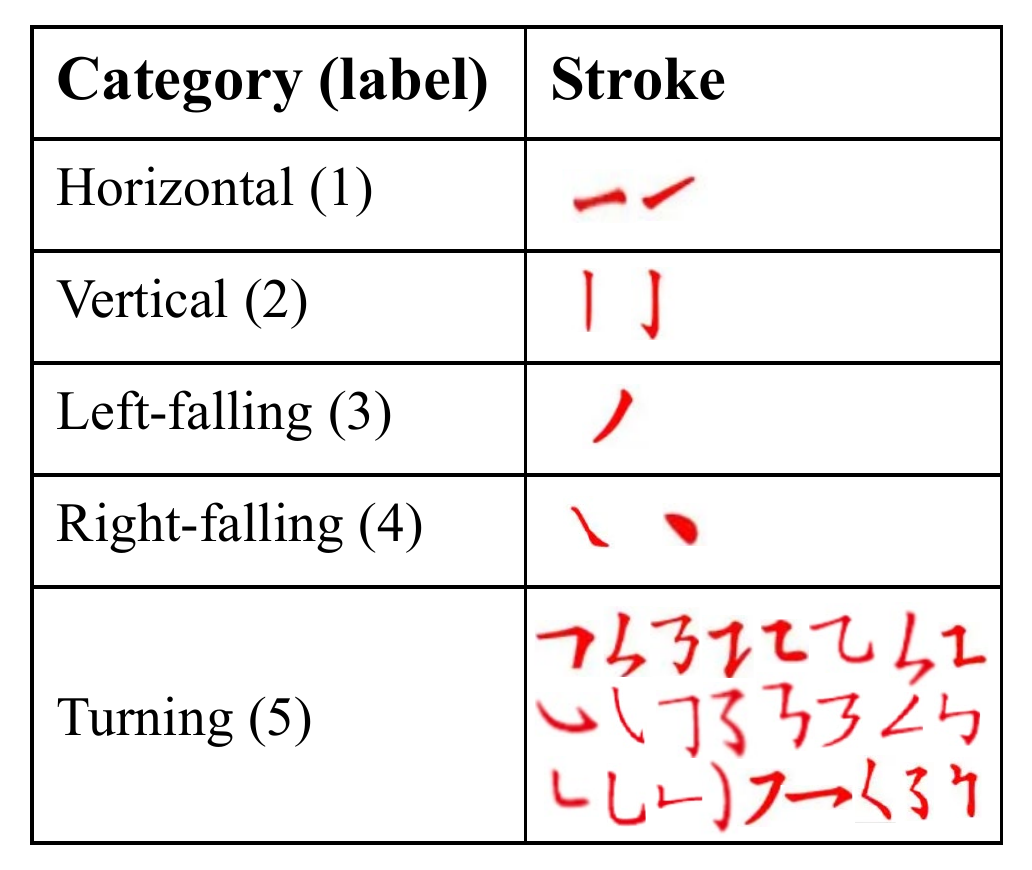}
  \begin{center}
\small (a) 5 basic categories of strokes.  
\end{center}  
 \end{minipage}
 \begin{minipage}{0.49\linewidth}
  \centering
  \includegraphics[width=.99\linewidth]{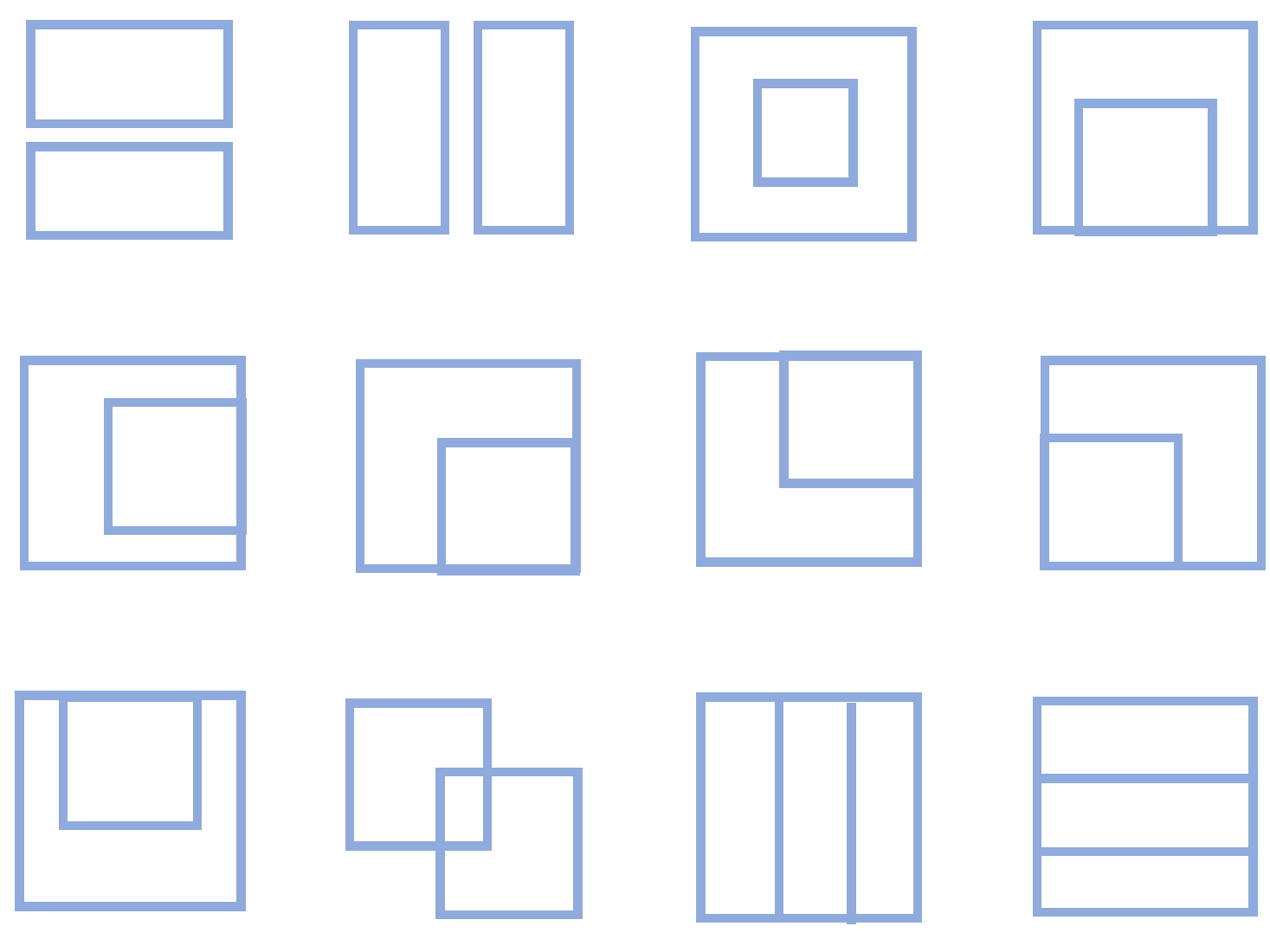}
    \begin{center}
\small (b) 12 basic spatial structures. 
\end{center} 
 \end{minipage}
 \begin{minipage}{1\linewidth}
  \centering
  \includegraphics[width=1\linewidth]{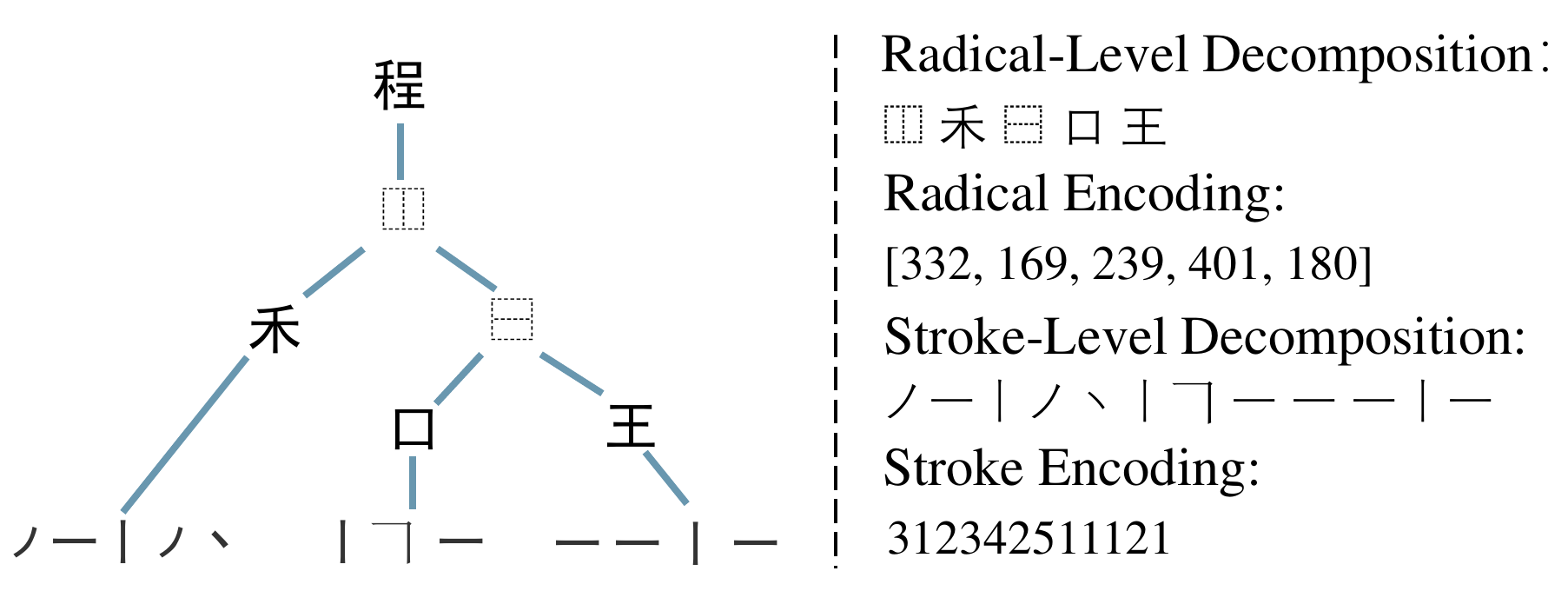}
  \centerline{{\small (c) Radical- and stroke-level decompositions of a Chinese character.}}
 \end{minipage}
  \caption{Strokes, spatial structures, and both stroke- and radical-level decompositions of Chinese characters.}
  \label{Fig: stroke_radical_encoding}
\end{figure*}

Existing zero-shot recognition methods can be generally divided into two categories, i.e., the radical-based methods \citep{DenseRAN_wang_2018,RAN_zhang_2018,FewShotRAN_wang_2019,HDE_cao_2020,REZCR_Diao_2022} and stroke-based methods \citep{SLD_chen_2021, SBA_chen_2022}. The main idea of the kind of radical-based approaches is to recognize unseen characters through learning the radical-level decomposition of Chinese characters, mainly motivated by the observation that each Chinese character is closely related to certain radical-level decomposition formed by a collection of radicals and spatial structures as shown in Figure \ref{Fig: stroke_radical_encoding}(b) and (c). In the early work \citep{DenseRAN_wang_2018} based on the radical-level decomposition, the authors proposed an effective method called DenseRAN, which uses an encoder-decoder network to learn a predefined two-dimensional radical-level decomposition. In \citep{HDE_cao_2020}, the authors suggested a novel hierarchical decomposition embedding (HDE) to represent the radical-level decomposition in the semantic level, and then identified Chinese characters by a vector matching scheme in the embedding space. The recent paper \citep{REZCR_Diao_2022} proposed an attention-based radical information extractor to yield the radical sequences and structural relationships representing the radical-level decompositions of Chinese characters, and then introduced a knowledge graph-based reasoner to infer the final result. Despite the impressive performance of this kind of radical-based methods in some benchmarks, they may suffer from the radical imbalance issue \citep{SLD_chen_2021}, that is, the occurrence frequencies of different radicals vary dramatically. Moreover, the kind of radical-based methods is hard to deal with the radical zero-shot scenario, i.e., existing some unseen radicals in the characters desired to recognize.

\begin{figure*}[ht]
\centering
\includegraphics[scale=0.8]{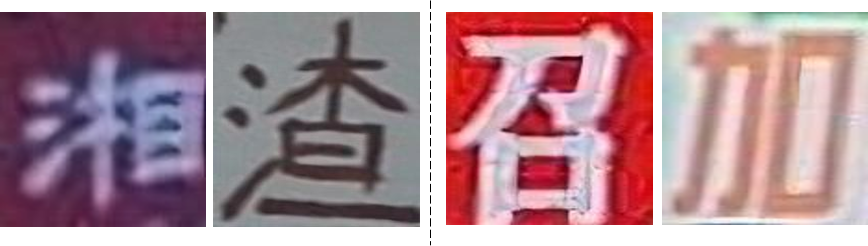}
\caption{Some characters with the same stroke-level decompositions but different radical-level decompositions.}
\label{Fig:similar-character}
\end{figure*}

To handle the radical zero-shot scenario, several stroke-based methods have been proposed in the recent literature. In \citep{SLD_chen_2021}, the authors utilized the stroke-level decomposition and adopted an encoder-decoder model to yield the stroke-level feature and stroke encoding in the training stage. To alleviate the similar stroke imbalance issue, \citep{SLD_chen_2021} divided the 32 basic strokes into five categories according to Figure \ref{Fig: stroke_radical_encoding}(a) motivated by the literature \citep{cw2vec_cao_2018}, and then suggested an effective stroke encoding based on these five major categories to represent the stroke-level decomposition as shown in Figure \ref{Fig: stroke_radical_encoding}(c). In \citep{SBA_chen_2022}, the authors proposed an encoder-decoder model for zero-shot Chinese character recognition by incorporating the stroke-level decomposition into the model. Although these stroke-based methods can achieve impressive performance in some important scenarios such as the recognition of printed artistic characters with well-defined stroke-level decomposition and can effectively handle the radical zero-shot setting, the performance of these methods is limited when applied to some difficult recognition tasks such as the recognition of characters with complex backgrounds. In particular, the sole stroke-level decomposition may be insufficient for zero-shot Chinese character recognition since there are some Chinese characters with the same stroke-level decompositions as depicted in Figure \ref{Fig:similar-character}.

Noticing that the stroke- and radical-level decompositions can provide different levels of information on Chinese characters as shown in Figure \ref{Fig: stroke_radical_encoding}, this paper suggests exploiting both of them for zero-shot Chinese character recognition. Our major contributions can be summarized as follows:
\begin{enumerate}
    \item[(1)] We propose an effective method for zero-shot Chinese character recognition based on \underline{ST}roke- \underline{A}nd \underline{R}adical-level decompositions (dubbed \textit{STAR}). The proposed method consists of a training stage and an inference stage. In the training stage, two similar encoder-decoder networks are adopted to yield respectively the estimates of stroke and radical encodings, which together with the true encodings are then utilized to formalize the associated stroke and radical losses for jointly training. A similarity loss is imposed to regularize both encoders to extract the features of the same Chinese characters with high correlation. 
    
    \item[(2)] In the inference stage, two key modules including the stroke screening module (SSM) and feature matching module (FMM) are proposed to deal with the deterministic and confusing cases respectively. In FMM, an effective stroke rectification scheme is proposed to yield an extended candidate set of characters for final inference, aiming to reduce the misdiagnosis rate.
    
    \item[(3)] A series of experiments over three benchmark datasets are conducted to show the effectiveness of our method. Numerical results show that the proposed method outperforms state-of-the-art methods in both character and radical zero-shot settings and maintains the competitive performance in the traditional seen character setting.
\end{enumerate}

The rest of this paper is organized as follows. In Section \ref{Related Work} we present some related work. In Section \ref{sc:proposed method}, we describe the proposed method in detail. In Section \ref{Experiments}, we conduct numerous experiments to demonstrate the effectiveness of the proposed method. We conclude this paper in Section \ref{Conclusion}.




\section{Related Work}
\label{Related Work}
Chinese character recognition has been extensively studied in the past decades \citep{feature_trier_1996,Normalization_liu_2003,similar_tao_2014}. Early work on Chinese character recognition usually relied on the hand-crafted features such as Gabor \citep{Gabor_Su_2003} or gradient feature maps \citep{gradient_feature_maps_chang_2006,gradient_feature_liu_2007}. The performance of these kinds of methods are limited by the low capabilities of these hand-crafted features when applied to some practical scenarios such as the recognition of handwritten Chinese characters \citep{Text_recognition_Chen_2021}. 

With the development of deep learning, especially the convolutional neural networks, many deep neural networks have been suggested in the literature to automatically extract features and achieved remarkable progress for the Chinese character recognition . The main idea of this kind of methods is to regard the Chinese character recognition as an image classification problem with multiple categories. Thus, these kinds of methods are called  the \textit{character-based} methods henceforth. In the seminal work  \citep{MCDNN_2013}, the authors proposed the first deep neural network model called MCDNN, achieving an accuracy rate close to that of humans when applied to Chinese character recognition. In \citep{HCCR_GoogLeNet_zhong_2015}, the authors adopted GoogLeNet \citep{GoogLeNet_szegedy_2015} to extract the latent features and integrated the traditional directional feature maps to improve the model performance. In \citep{DirectMap_zhang_2017}, the authors combined the convolutional neural networks with the direction-decomposed feature map, and reduced the mismatch between training and test data on specific source layers by adapting layers to improve the performance. \citep{M-RBC+IR_Xiao_2017} suggested an iterative refinement module applied to the output feature maps of convolutional neural networks to alleviate the confusion problem of morphologically similar characters. \citep{Template_Instance_xiao_2019} proposed a template loss and an instance loss to deal with small differences in morphological close characters. Despite the great success of this kind of character-based methods, they generally rely on a large amount of training data, of which the collection is expensive and labor-cost. Thus, these methods are not applicable to the few/zero-shot Chinese character recognition \citep{FewShotRAN_wang_2019}.

Distinguished from the above methods, this paper focuses on the zero-shot setting, which is involved in many real-word applications such as autonomous driving \citep{driving_zhang_2021,drive_qian_2015} and historical documents retrieval \citep{document_maekawa_2019,obc306_huang_2019}. Motivated by the observation that the stroke- and radical-level decompositions can provide different levels of information, we propose a novel zero-shot Chinese character recognition method by integrating them. Our idea is also very different to existing works on zero-shot Chinese character recognition \citep{DenseRAN_wang_2018,FewShotRAN_wang_2019,HDE_cao_2020,REZCR_Diao_2022,SLD_chen_2021,SBA_chen_2022}, which exploit either the radical-level or stroke-level decomposition, as discussed before.

\begin{figure*}[t]
\centering
\includegraphics[scale=0.49]{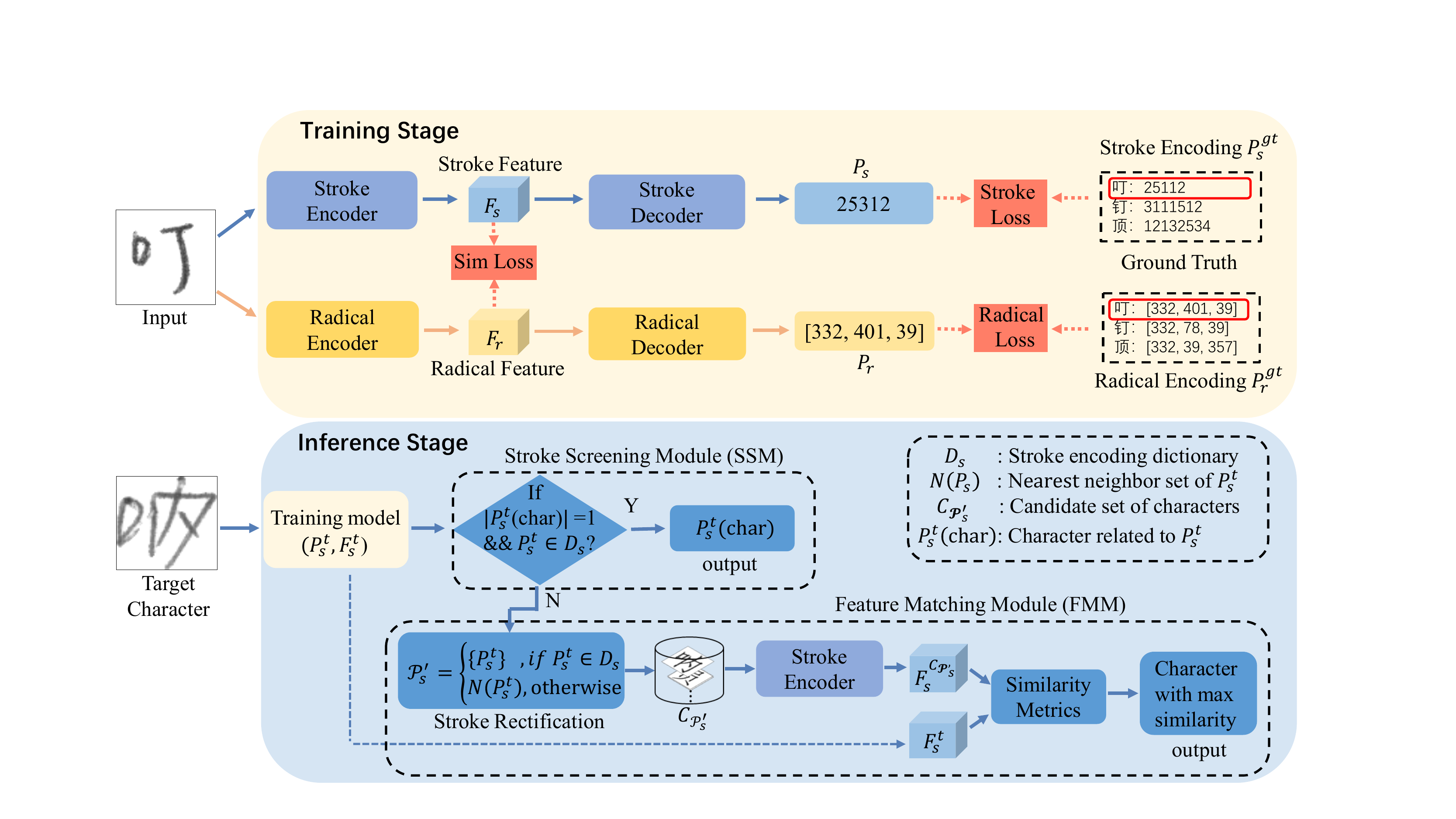}
\caption{The architecture of the proposed STAR model, consisting of a training stage and an inference stage.}
\label{Fig:model}
\end{figure*}

\section{Proposed Method: STAR}
\label{sc:proposed method}

In this section, we describe the proposed method for zero-shot Chinese character recognition in detail. The core idea of the proposed method is to exploit both stroke- and radical-level decompositions of Chinese characters through a joint training way, mainly motivated by the fact that the stroke- and radical-level decompositions can provide different levels of important information for recognizing Chinese characters. In the following, we first introduce the stroke and radical encodings used in this paper to represent the associated stroke- and radical-level decompositions, and then describe the architecture of the proposed method and its training loss.

\subsection{Stroke and Radical Encodings}
\label{Stroke and Radical Encodings}
As shown in Figure \ref{Fig: stroke_radical_encoding}, all Chinese characters can be uniquely decomposed into strokes or radicals and spatial structures. According to the Chinese national standard GB18030-2005, all Chinese characters are composed of 32 basic strokes, which can be divided into five major categories \citep{cw2vec_cao_2018,SLD_chen_2021}, namely horizontal, vertical, left-falling, right-falling, and turning as shown in Figure \ref{Fig: stroke_radical_encoding}(a). Thus, for each Chinese character, we can use a simple encoding of the dimension being its number of strokes to encode its stroke-level decomposition as done in \citep{SLD_chen_2021}. Specifically, the $i$-th component of the stroke encoding is $j \in \{1,2,3,4,5\}$ if the $i$-th stroke of this character belongs to the $j$-th category presented in Figure \ref{Fig: stroke_radical_encoding}(a). A specific example is shown in Figure \ref{Fig: stroke_radical_encoding}(c), where the stroke encoding of the concerned character is ``312342511121''.

Then we turn to define the radical encoding used in this paper. Similarly, according to the Chinese national standard GB13000.1, there are 394 radicals and 12 basic spatial structures which are composed of radical encoding for the 3,755 commonly used Chinese characters as shown in Figure \ref{Fig: stroke_radical_encoding}(b). Based on the database CJKVI-IDS, each Chinese character can be uniquely decomposed into an ideographic description sequence (IDS) formed by these radicals and spatial structures. Thus, we can define the radical encoding of a Chinese character according to its IDS as the radical-level decomposition, through a similar defining way of the stroke encoding. A specific example of the radical encoding is presented in Figure \ref{Fig: stroke_radical_encoding}(c), where there are five components in the radical-level decomposition of the concerned character and thus its radical encoding contains five values by sequence, representing the labels of these five components in the radical-level decomposition respectively.

\subsection{Architecture and Training Loss}
\label{Architecture and Training Loss}
The architecture of the proposed method is depicted in Figure \ref{Fig:model}. The proposed method includes a training stage and a inference stage. In the training stage, two similar encoder-decoder networks are adopted respectively to extract the features in the stroke and radical levels and also yield the stroke and radical encodings, which together with the true encodings are used to formalize the stroke loss and radical loss for jointly training. Besides the stroke loss and radical loss, we introduce a similarity loss between these two encoder-decoder models to regularize both encoders to extract features of the same characters with high correlation. Specifically, the stroke loss $L_{stroke}$, radical loss $L_{radical}$ and similarity loss $L_{sim}$ are defined as follows:
\begin{align*}
   L_{stroke} = CE(P_s, P_s^{gt}), \ L_{radical} = CE(P_r, P_r^{gt}),\
   L_{sim} = 1 - \frac{F_s^TF_r}{\|F_s\|\cdot \|F_r\|},
\end{align*}
where $P_s$ ($P_r$) and $P_s^{gt}$ ($P_r^{gt}$) represent the predicted and true stroke (radical) encodings respectively, $CE(\cdot)$ represents the cross entropy loss, $F_s$ and $F_r$ are the features extracted by the stroke and radical encoders respectively. Thus, the total training loss can be formulated as
\begin{align*}
   L_{star} = L_{stroke} +  L_{radical} + \lambda L_{sim},
\end{align*}
where $\lambda$ is a tuning hyperparameter.

In the inference stage, there are two key modules, i.e., the stroke screening module (SSM) and feature matching module (FMM)  designed to tackle the deterministic and confusing cases, respectively. Since the stroke-level information is a kind of fine-grained information and generally more suitable than the radical-level information for the inference in zero-shot Chinese character recognition, we prefer to use the predicted stroke encoding in the inference stage as done in the literature \citep{SLD_chen_2021}. The specific workflow at the inference stage can be described as follows: given an image of a target Chinese character desired to recognize, we at first feed it into the training model trained in advance to yield the predicted stroke encoding $P_s^t$ as well as its stroke feature $F_s^{t}$, then output the character $P_s^t(char)$ related to $P_s^t$ if the predicted stroke encoding $P_s^t$ is in the predefined stroke encoding dictionary $D_s$ and also is related to a unique Chinese character, and  otherwise turn to FMM. In this case, it is hard to directly infer the character since $P_s^t$ is either not in $D_s$ or related to multiple characters. To deal with these two confusing cases in a unified way, we introduce an effective stroke rectification scheme in FMM to yield an effective candidate set of characters for the final inference. Specifically, we define a rectified stroke encoding set ${\cal P}_s'$ as $P_s^t$ itself if it is in $D_s$ and the nearest neighbor set $N(P_s^t) \subset D_s$ of $P_s^t$ otherwise. Based on the rectified stroke encoding set ${\cal P}_s'$, we yield a candidate set of characters ${\cal C}_{{\cal P}_s'}$ for the final inference, through collecting all the Chinese characters related to stroke encodings in ${\cal P}_s'$. We then feed the images related to these characters in ${\cal C}_{{\cal P}_s'}$ from a preprepared support sample set to the stroke encoder trained in the training stage to yield their stroke feature set $F_s^{{\cal C}_{{\cal P}_s'}}$, and later match the similarity between $F_s^{{\cal C}_{{\cal P}_s'}}$ and $F_s^{t}$, and finally output the character with the maximal similarity as the eventual inference result.   

Distinguished to the previous methods suggested in the literature, the proposed method exploits both the stroke- and radical-level decompositions in the training stage due to different level information provided by these two decompositions, while existing methods utilize either the radical-level decomposition \citep{DenseRAN_wang_2018,FewShotRAN_wang_2019,HDE_cao_2020,REZCR_Diao_2022} or the stroke-level decomposition \citep{SLD_chen_2021,SBA_chen_2022}. Moreover, as compared to the inference stages of existing stroke-level decomposition based methods \citep{SLD_chen_2021,SBA_chen_2022}, the proposed method uses the nearest neighbor set $N(P_s^t)$ including all active stroke encodings in $D_s$ with the smallest Levenshtein distance to the inactive predicted stroke encoding $P_s^t$ in the stroke rectification scheme, aiming to reduce the misdiagnosis rate by expanding the candidate set, while existing methods in \citep{SLD_chen_2021,SBA_chen_2022} only use one of the nearest stroke encodings as the rectified stroke encoding (generally taking the first one in $D_s$). As shown in the later Table \ref{Table:ablation} and Figure \ref{Fig:test_example}, the suggested stroke rectification scheme is more effective than existing ones \citep{SLD_chen_2021,SBA_chen_2022}.

\section{Experiments}
\label{Experiments}
In this section, we conducted numerous experiments to demonstrate the effectiveness of the proposed method in comparison to the state-of-the-art methods. All experiments were carried out in Pytorch environment running Linux, NVIDIA RTX 3090 Ti GPU.


\subsection{Experimental Settings}
\label{Experimental Settings}
We first describe the specific datasets used in this paper, and then present the specific network architectures of the proposed model and finally point out the optimizer used in this paper as well as some hyperparameter settings.

\subsubsection{Datasets}
\label{Dataset}
We considered the following three datasets.
\begin{itemize}
    \item The handwritten dataset was collected from the well-known CASIA-HWDB 1.0-1.1  \citep{HWDB_liu_2013} and ICDAR2013 \citep{ICDAR_Yin_2013} databases,  where HWDB1.0-1.1 contains 2,678,424 offline images of handwritten Chinese characters divided into 3,881 classes and collected by 720 persons, and ICDAR2013 contains 224,419 offline handwritten Chinese character images, divided into 3,755 classes and collected by 60 persons.
    
    \item The printed artistic dataset was released by \citep{SLD_chen_2021}. It consists of 394,275 samples in total, formed by 105 typographic art fonts and divided into 3,755 commonly used Chinese character categories.
    
    \item The CTW dataset was released by \citep{CTW_yuan_2019}. It is a dataset of Chinese texts with about 1 million Chinese characters from 3,850 unique ones annotated by experts in over 30,000 street view images. Specifically, there are 32,285 high resolution street view images, 1,018,402 character instances, 3,850 character categories and 6 kinds of attributes in the CTW dataset. 
\end{itemize}

\subsubsection{Network Architectures}
\label{Network Architectures}
The stroke and radical encoders in the proposed method are the same as that in \citep{SLD_chen_2021}, consisting of two convolutional layers, one max-pooling layer, and 16 residual blocks. The stroke and radical decoders are consistent, both using the original Transformer decoder \citep{transformer_vaswani_2017}.

\subsubsection{Optimizer}
\label{Optimizer}
We used the popular Adadelta \citep{adadelta_zeiler_2012}  as the optimizer with the learning rate $1$. The hyperparameter $\lambda$ was empirically set as 0.1 according to the later ablation study. The batch size was set as 32. Each input image was resized to 32 $\times32$ and normalized into [-1,1]. We used the recognition accuracy as the evaluation metric.

\begin{table}[ht]\footnotesize
\caption{Comparison results on the recognition accuracy (\%) of different methods in the character zero-shot settings. The best results of the state-of-the-art (SOTA) models are marked in blue color. The improved ratio is defined as $(\frac{\text{performance of STAR}}{\text{performance of SOTA}}-1)\times 100\%$.}

\centering
\scriptsize
\centering\renewcommand\arraystretch{1.3}
\setlength{\tabcolsep}{4.5mm}{
\begin{tabular}{l|ccccc}
\bottomrule
\multirow{2}{*}{Handwritten Dataset} & \multicolumn{5}{c}{$m$ for Character Zero-Shot Setting } \\ \cline{2-6}
& 500     & 1000    & 1500    & 2000    & 2755  \\ \cline{1-6}
DenseRAN \citep{DenseRAN_wang_2018}   & 1.70  & 8.44  & 14.71  & 19.51  & 30.68        \\
HDE \citep{HDE_cao_2020} & 4.90  & 12.77 & 19.25  & 25.13  & 33.49    \\
SLD \citep{SLD_chen_2021}  & 2.94  & 11.53 & 21.90  & 28.90  & 36.04       \\
SBA \citep{SBA_chen_2022} & \textcolor{blue}{5.91}  & \textcolor{blue}{14.35} & \textcolor{blue}{24.32}  & \textcolor{blue}{30.17}  & \textcolor{blue}{40.22}   \\
STAR (this paper)  &\textbf{7.54}   & \textbf{19.47} & \textbf{27.79} & \textbf{35.53} & \textbf{43.86}  \\\cline{1-6}
Improved ratio (\%)  & 27.58  & 35.68  & 14.27  & 17.77 & 9.05\\
\bottomrule
\multirow{2}{*}{Printed Artistic Dataset} & \multicolumn{5}{c}{$m$ for Character Zero-Shot Setting}   \\ \cline{2-6}
& 500     & 1000    & 1500    & 2000    & 2755     \\ \cline{1-6}
DenseRAN \citep{DenseRAN_wang_2018}  & 0.20   & 2.26    & 7.89    & 10.86   & 24.80   \\
HDE \citep{HDE_cao_2020}             & 7.48   & 21.13   & 31.75   & 40.43   & 51.41   \\
SLD \citep{SLD_chen_2021}            & 4.57   & \textcolor{blue}{34.25}   & \textcolor{blue}{51.98}   & \textcolor{blue}{60.67}   & \textcolor{blue}{69.04}   \\
SBA \citep{SBA_chen_2022}            & \textcolor{blue}{8.25}   & 32.24   & 50.13   & 57.13   & 68.88   \\
STAR (this paper)                   & \textbf{16.42}  & \textbf{50.02}  & \textbf{65.94} & \textbf{73.54}  & \textbf{80.00}      \\\cline{1-6}
Improved ratio (\%) &99.03  &46.04  &26.86  &21.21 & 15.87\\
\bottomrule
\multirow{2}{*}{CTW Dataset} & \multicolumn{5}{c}{$m$ for Character Zero-Shot Setting}  \\ \cline{2-6}
& 500     & 1000    & 1500    & 2000    & 2241 \\ \cline{1-6}
SLD \citep{SLD_chen_2021}    & \textcolor{blue}{0.60}  & \textcolor{blue}{2.78}  & \textcolor{blue}{4.89}  & \textcolor{blue}{7.80}  & \textcolor{blue}{10.30}       \\
STAR (this paper)  & \textbf{1.19}  & \textbf{3.77}  & \textbf{8.04}  & \textbf{11.00} & \textbf{11.27}  \\\cline{1-6}
Improved ratio (\%) &98.33  &35.61  &64.42  &41.03 & 9.97\\
\bottomrule
\end{tabular}}
\label{Table:character_zero-shot}
\end{table}

\subsection{Experimental Results}
\label{Experimental Results}
We evaluated the performance of the proposed method in the following three scenarios, i.e., the character zero-shot setting, the radical zero-shot setting, and the traditional seen character setting.
\begin{itemize}
    \item In the character zero-shot setting, the characters desired to recognize are not in the training set.
    
    \item In the radical zero-shot setting, the radicals of characters desired to recognize are not in the training set.
    
    \item In the traditional seen character setting, the data is split into the training and test sets by a regular way.
\end{itemize}

\begin{table} [ht]
\caption{Comparison results on the recognition accuracy (\%) of different methods in the radical zero-shot settings. The best results of the state-of-the-art models are marked in blue color.}
\centering
\scriptsize
\centering\renewcommand\arraystretch{1.3}
\setlength{\tabcolsep}{4.5mm}{
\begin{tabular}{l|ccccc}
\bottomrule
\multirow{2}{*}{Handwritten Dataset} & \multicolumn{5}{c}{$n$ for Radical Zero-Shot Setting} \\ \cline{2-6}
& $\geq$50    & $\geq$40    & $\geq$30   & $\geq$20    & $\geq$10  \\
\cline{1-6}
DenseRAN \citep{DenseRAN_wang_2018}  & 0.21  & 0.29  & 0.25  & 0.42  & 0.69 \\
HDE \citep{HDE_cao_2020} & 3.26  & 4.29  & 6.33   & 7.64   & 9.33 \\
SLD \citep{SLD_chen_2021}   & \textcolor{blue}{5.28}  & \textcolor{blue}{6.87} & \textcolor{blue}{9.02}  & \textcolor{blue}{15.67}   & \textcolor{blue}{16.31}  \\
STAR (this paper) & \textbf{6.95}  & \textbf{12.28} & \textbf{14.74} & \textbf{18.37} & \textbf{23.23} \\ \hline
Improved ratio (\%) &31.63 &78.75 &63.41  &17.23  &42.43\\
\bottomrule
\multirow{2}{*}{Printed Artistic Dataset} & \multicolumn{5}{c}{$n$ for Radical Zero-Shot Setting} \\ \cline{2-6}
& $\geq$50    & $\geq$40    & $\geq$30   & $\geq$20    & $\geq$10  \\
\cline{1-6}
DenseRAN \citep{DenseRAN_wang_2018}  & 0.07  & 0.16  & 0.25  & 0.78  & 1.15 \\
HDE \citep{HDE_cao_2020}  & 4.85  & 6.27  & 10.02   & 12.75   & 15.25  \\
SLD \citep{SLD_chen_2021}     & \textcolor{blue}{10.10}  & \textcolor{blue}{18.17} & \textcolor{blue}{24.34}  & \textcolor{blue}{36.52}   & \textcolor{blue}{40.78} \\
STAR (this paper) & \textbf{17.46}  & \textbf{21.63} & \textbf{35.31} & \textbf{42.58} & \textbf{48.10} \\\hline
Improved ratio (\%) &72.87  & 19.04 & 45.07  & 16.59 & 17.95\\
\bottomrule
\multirow{2}{*}{CTW Dataset} & \multicolumn{5}{c}{$n$ for Radical Zero-Shot Setting}  \\ \cline{2-6}
& $\geq$50    & $\geq$40    & $\geq$30   & $\geq$20    & $\geq$10  \\ \cline{1-6}
SLD \citep{SLD_chen_2021}    & \textcolor{blue}{0.66}  & \textcolor{blue}{0.78}  & \textcolor{blue}{0.86}  & \textcolor{blue}{1.21}  & \textcolor{blue}{2.82}        \\
STAR (this paper)  & \textbf{2.16}  & \textbf{2.33}  & \textbf{2.76}  & \textbf{4.81} & \textbf{5.35}  \\ \hline
Improved ratio (\%) &227.27 &198.72 &220.93  &297.52  &89.72\\
\bottomrule
\end{tabular}}
\label{Table:Radical_zero-shot}
\end{table}

\subsubsection{Experiments in Character Zero-shot Setting}
\label{Experiments in Character Zero-shot Setting}
We implemented a series of experiments over three benchmark datasets described before. For the \textbf{handwritten} dataset,  we split it according to the same way as in \citep{SLD_chen_2021} for a fair comparison. Specifically, we first selected the samples ranked in the top $m$ categories of the 3,755 commonly used Chinese character categories from HWDB1.0-1.1 as the training set, where $m$ ranges from $ \{500, 1000, 1500, 2000, 2755\} $, then collected the Chinese characters ranked in the last 1,000 categories from ICDAR2013 as the test set. This ensures that the character categories and handwritten styles of both the training and test sets do not overlap. For the \textbf{printed artistic} dataset, we used the similar data splitting way to yield the training and test sets from on their own. 
For the \textbf{CTW} dataset, since the test set was not released publicly, we only used the released training and validation datasets in our experiments. According to the provided annotations, we first cropped 537,236 images of individual Chinese characters from the training and validation sets. Due to the low quality of images, we used some image enhancement methods such as SRGAN \citep{SRGAN_Ledig_2017} to improve the resolution of images, and then filtered out the images with Laplace operator edge blurring degree value less than 50, and finally yielded 362,495 single-character images with 2741 classes. Based on the data after pre-processing, we considered five character zero-shot cases. Similarly, we selected the samples ranked in the top $m$ categories of the 2,741 commonly used Chinese character categories as the training set, where $m$ ranges from $ \{500, 1000, 1500, 2000, 2241\} $, and then selected Chinese characters ranked in the last 500 categories as the test set.

To demonstrate the effectiveness of the proposed method, we considered four state-of-the-art methods including two radical-based methods namely, DenseRAN \citep{DenseRAN_wang_2018} and HDE \citep{HDE_cao_2020}, and two stroke-based methods i.e., SLD \citep{SLD_chen_2021} and SBA \citep{SBA_chen_2022} as baselines for the handwritten and printed artistic datasets, and only SLD reproduced by ourselves as the baseline for CTW, since we cannot access the reproducible codes of the other three methods. The comparison results on the recognition accuracy are presented in  
Table \ref{Table:character_zero-shot}.

From Table \ref{Table:character_zero-shot}, the proposed method significantly outperforms existing methods exploiting either radical-level decomposition (i.e., DeseRAN \citep{DenseRAN_wang_2018} and HDE \citep{HDE_cao_2020}) or stroke-level decomposition (i.e., SLD \citep{SLD_chen_2021} and SBA \citep{SBA_chen_2022}) in all five concerned character zero-shot settings. Specifically, the improved ratios\footnote{The improved ratio is defined as $(\frac{\text{performance of STAR}}{\text{performance of SOTA}}-1)\times 100\%$.} over the handwritten dataset for five different character zero-shot settings are 27.58\%, 35.68\%, 14.27\%, 17.77\% and 9.05\%, respectively, and the improvements of the proposed method are more significantly over both the printed artistic dataset and CTW dataset than the handwritten dataset, with about twice of the state-of-the-art recognition accuracy in the character zero-shot setting with only 500 categories for training over both the printed artistic dataset and CTW dataset. Moreover, the improvement achieved by the proposed method is generally more significant in the zero-shot setting with fewer categories of Chinese characters for training. These show clearly that the effectiveness of the proposed method by exploiting both stroke- and radical-level decompositions to capture different-level important information for zero-shot Chinese character recognition. 

When regarding the performance over different datasets, it can be observed from Table \ref{Table:character_zero-shot} that all methods achieve better performance over both the handwritten dataset and printed artistic dataset than the CTW dataset, since the CTW dataset is a more challenging dataset containing a variety of text images such as the under-lit and partially occluded text images.

\subsubsection{Experiments in Radical Zero-Shot Setting}
\label{Experiments in Radical Zero-Shot Setting}
In the following, we describe the experimental results in the radical zero-shot setting. Different from the character zero-shot setting, we divided the dataset according to the emerging frequencies of radicals in these 3,755 commonly used Chinese characters in the radical zero-shot setting. Specifically, we collected all the characters with the radicals whose emerging frequencies are more than $n$ times (where $ n\in \{50,40,30,20,10\} $) as the training set, and the rest characters as the test set. It should be pointed out that for the handwritten dataset, the training set was selected from CASIA-HWDB1.0-1.1 and the test set was selected from ICDAR 2013.
The comparison results are presented in Table \ref{Table:Radical_zero-shot}. 

From Table \ref{Table:Radical_zero-shot}, the proposed method outperforms existing methods over all radical zero-shot settings with significant improvements on the recognition accuracy. Specifically, the improved ratios on the recognition accuracy of the proposed method over the handwritten dataset are 31.63\%, 78.75\%, 63.41\%, 17.23\% and 42.43\% for five different radical zero-shot settings respectively. It can be observed that the proposed method achieve the greatest improvement in the radical zero-shot setting with $n=50$ over the handwritten dataset. Similar claim can be concluded for experimental results over the printed artistic dataset, where the proposed method achieve the greatest improvement in the radical zero-shot setting with $n=50$. When particularly concerning the CTW dataset, the proposed method generally achieves about twice improvements on the recognition accuracy in the cases $n=50, 40, 30$ and 20, and about once improvement in the case $n=10$, as compared to the state-of-the-art method SLD \citep{SLD_chen_2021}. These demonstrate the effectiveness of the proposed method and show clearly that both kinds of stroke and radical information are very important for zero-shot Chinese character recognition. 

\begin{table} [ht]
\caption{Comparison results of different methods over the handwritten dataset in the seen character setting.}
\centering
\footnotesize
\centering\renewcommand\arraystretch{1.5}
\setlength{\tabcolsep}{4.5mm}{
\begin{tabular}{lc} \hline
Method & Accuracy (\%)          \\ \hline
Human Performance \citep{ICDAR_Yin_2013}            & 96.13          \\
HCCR-GoogLeNet (Zhong et al. 2015) & 96.35          \\
DirectMap+ConvNet+Adaptation (Zhang et al.2017) & 97.37           \\
M-RBC+IR \citep{M-RBC+IR_Xiao_2017}  & 97.37           \\
Template+Instance\citep{Template_Instance_xiao_2019}            & \textbf{97.45}  \\ \hline
DenseRAN \citep{DenseRAN_wang_2018}                     & 96.66           \\
FewShotRAN\citep{FewShotRAN_wang_2019}                   & 96.97           \\
HDE \citep{HDE_cao_2020}                          & 96.74           \\
SLD \citep{SLD_chen_2021}         & 96.28           \\
REZCR \citep{REZCR_Diao_2022}         & 96.35          \\
STAR (this paper)                         & \textbf{97.11}
\\ \hline
\end{tabular}}
\label{Table:ICDAR2013}
\end{table}

\begin{table*}[ht]
\scriptsize
\caption{Ablation study results over the printed artistic dataset in the character zero-shot settings with five different numbers of categories $m$ varying from $\{500, 1000, 1500, 2000, 2755\}$, where the recognition accuracy (\%) was recorded.}
\centering\renewcommand\arraystretch{1.0}
\setlength{\tabcolsep}{0.9mm}{
\begin{tabular}{l|cc|cc|cc|cc|cc} 
\bottomrule
\multirow{2}{*}{Training Model} & \multicolumn{2}{c|}{500} & \multicolumn{2}{c|}{1000} & \multicolumn{2}{c|}{1500} & \multicolumn{2}{c|}{2000} & \multicolumn{2}{c}{2755}  \\ 
\cline{2-11}
& Infer$_{\text{1st}}$ & Infer$_{\text{all}}$    & Infer$_{\text{1st}}$ & Infer$_{\text{all}}$    & Infer$_{\text{1st}}$ & Infer$_{\text{all}}$    & Infer$_{\text{1st}}$ & Infer$_{\text{all}}$   & Infer$_{\text{1st}}$ & Infer$_{\text{all}}$             \\ 
\hline
Stroke          & 4.57   & 8.93    & 34.35  & 47.27       & 51.98  & 62.47       & 60.67  & 70.53      & 69.04  & 76.66        \\
Stroke + Radical   & 8.29   & \textbf{16.42}      & 37.48  & \textbf{50.02}       & 55.46  & \textbf{65.94}       & 64.42  & \textbf{73.54}       & 72.84  & \textbf{80.00}       \\
\bottomrule
\end{tabular}}
\label{Table:ablation}
\end{table*}

\begin{figure*}[ht]
\centering
\includegraphics[scale=0.395]{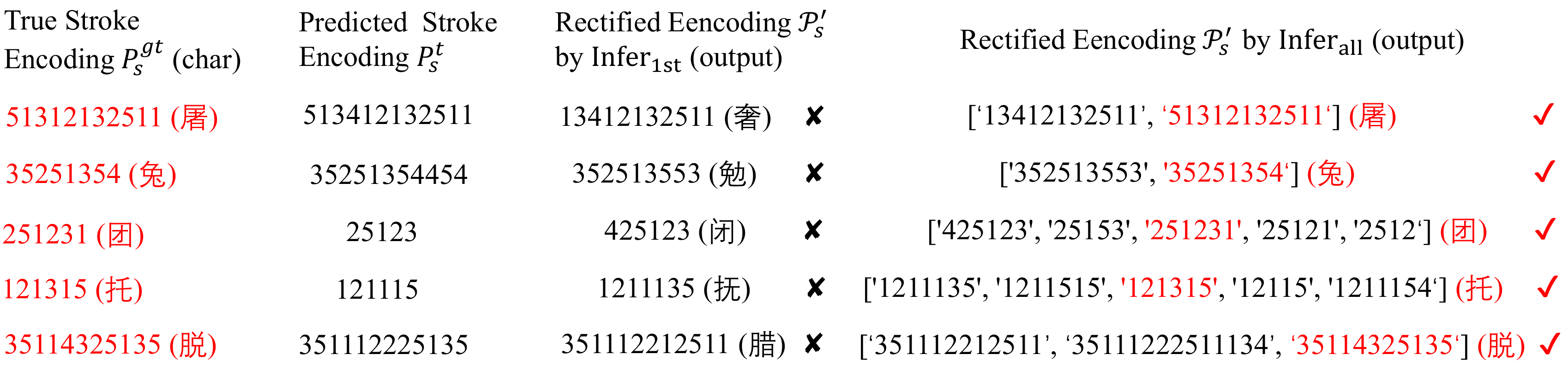}
\caption{Some test examples of the suggested inference model \textbf{Infer$_{\text{all}}$} and existing inference model \textbf{Infer$_{\text{1st}}$} \citep{SLD_chen_2021}. It can be observed that the suggested inference model can yield more accurate results by enlarging the candidate set.}
\label{Fig:test_example}
\end{figure*}

\subsubsection{Experiments in Seen Character Setting}
\label{Experiments in Seen Character Setting}
In this subsection, we considered the performance of the proposed method in the seen character setting over the handwritten dataset. We trained the model over CASIA-HWDB1.0-1.1 dataset and tested the model over the competition dataset ICDAR2013. The comparison results are presented in Table \ref{Table:ICDAR2013}. 

The first block of Table \ref{Table:ICDAR2013} presents the performance of the kind of character-based methods as well as human-beings. The second block of Table \ref{Table:ICDAR2013} presents the performance of the kind of few/zero-shot methods, where DenseRAN, FewShotRAN, HDE, REZCR are based on the radical-level decomposition and SLD is based on the stroke-level decomposition, and FewShotRAN is proposed for the few-shot setting while the rest four methods are proposed for the zero-shot setting. It can be observed that the proposed method still outperforms existing few/zero-shot methods and maintains the competitive performance as compared to the state-of-the-art results as well as human performance. These demonstrate the effectiveness of the proposed method.

\subsection{Ablation study}
\label{Ablation study}
We conducted a series of ablation studies to verify the feasibility of our idea and determine the hyperparameter $\lambda$. 

\textbf{A. On feasibility of the proposed idea.}
To verify our idea, we considered one training models and two inference models, described as follows:
\begin{itemize}
    \item \textbf{Stroke} represents the training model using only the stroke branch of the proposed training model.
    
    
    \item \textbf{Stroke + Radical} represents the proposed training model using both stroke and radical branches and the similarity loss.
    
    \item \textbf{Infer$_{\text{1st}}$} represents the inference model using only the first one of the nearest rectified stroke encodings in FMM as suggested in \citep{SLD_chen_2021}.
    
    \item \textbf{Infer$_{\text{all}}$} represents the proposed inference model using all of the nearest rectified stroke encodings in FMM. 
\end{itemize}
It should be pointed out that the training model \textbf{Stroke} together with the inference model \textbf{Infer$_{\text{1st}}$} is equal to the existing method SLD \citep{SLD_chen_2021}, while the proposed model is equal to \textbf{Stroke + Radical} together with  \textbf{Infer$_{\text{all}}$}. For each model, we implemented a set of experiments over the printed artistic dataset in character zero-shot settings with different $m$ varying from $\{500, 1000, 1500, 2000, 2755\}$. The experiment results on recognition accuracy are presented in Table \ref{Table:ablation}. 

\begin{figure*}[h]
\centering
\includegraphics[scale=0.49]{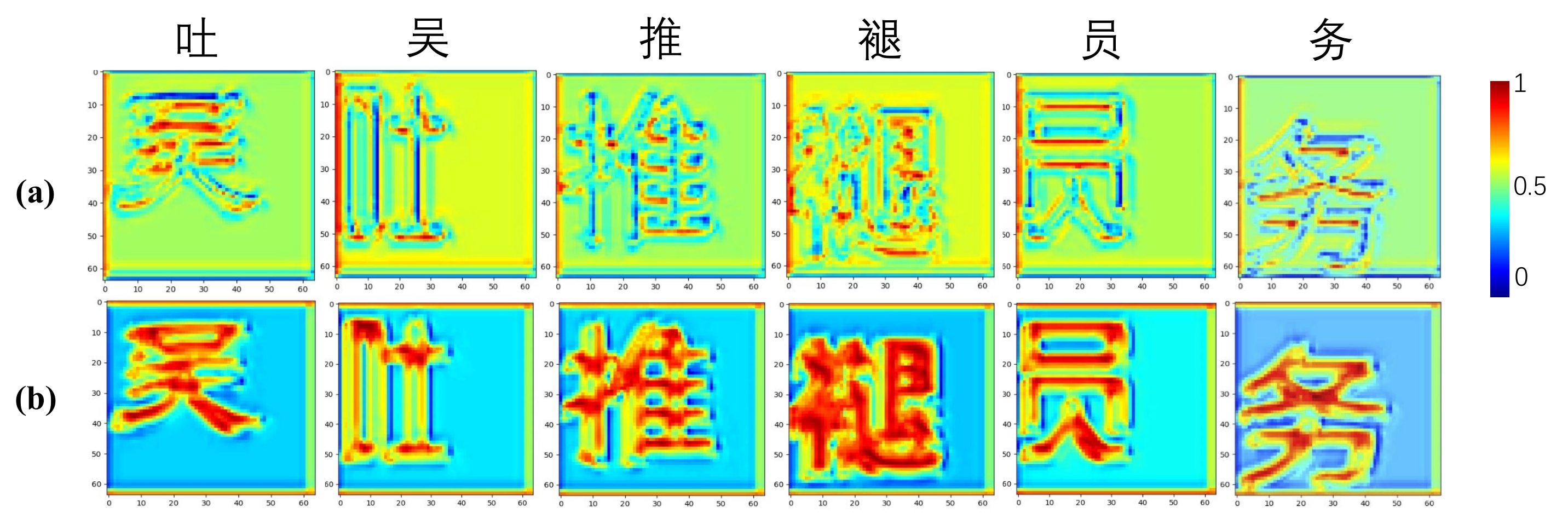}
\caption{Some visualization results of features extracted by the concerned two training models: (a) \textbf{Stroke}; (b) \textbf{Stroke+Radical}. It can observed that the suggested training model \textbf{Stroke+Radical} can capture the features of Chinese characters more accurately, through using both stroke- and radical-level decompositions.}
\label{Fig:visualization-feature}
\end{figure*}

Regarding the training model, it can be observed from Table \ref{Table:ablation} that the suggested training model \textbf{Stroke + Radical} using both stroke- and radical-level decompositions achieves better performance than the training model \textbf{Stroke} using only the stroke-level decomposition in all experimental settings. Particularly, in the case of $m=500$, the improvement achieved by the training model \textbf{Stroke + Radical} is more significant than the other four cases. We also present some visualization results of features extracted by the concerned two training models in Figure \ref{Fig:visualization-feature}. It can be observed from Figure  \ref{Fig:visualization-feature} that the training model \textbf{Stroke + Radical} using both stroke- and radical-level decompositions can pay more accurate attention to the characters than the training model \textbf{Stroke} using only the stroke-level decomposition. These show clearly that the effectiveness of the proposed training model, and in particular, the use of both stroke- and radical-level decompositions can provide more information for the zero-shot Chinese character recognition.

Regarding the inference model, it can be observed from Table \ref{Table:ablation} that the suggested inference model \textbf{Infer$_{\text{all}}$} using all the nearest rectified stroke encodings significantly outperforms the inference model \textbf{Infer$_{\text{1st}}$} suggested in the literature \citep{SLD_chen_2021} using the first one of the nearest rectified stroke encodings, mainly due to the enlarged candidate set of characters. Some test examples are presented in Figure \ref{Fig:test_example}. From Figure  \ref{Fig:test_example}, we can observe that the true stroke encodings of some characters are not the first ones of their nearest rectified stroke encodings and thus will be recognized incorrectly by the inference model \textbf{Infer$_{\text{1st}}$} as suggested in \citep{SLD_chen_2021}, while the suggested inference model \textbf{Infer$_{\text{all}}$} can effectively reduce the misdiagnosis rate by enlarging the candidate set. These demonstrate the effectiveness of the proposed inference model, in particular the suggested stroke rectification scheme.

\begin{figure*}[ht]
    \setlength{\abovecaptionskip}{-10pt} \setlength{\belowcaptionskip}{-1pt}
    \begin{center}
        \includegraphics[width=0.9\textwidth]{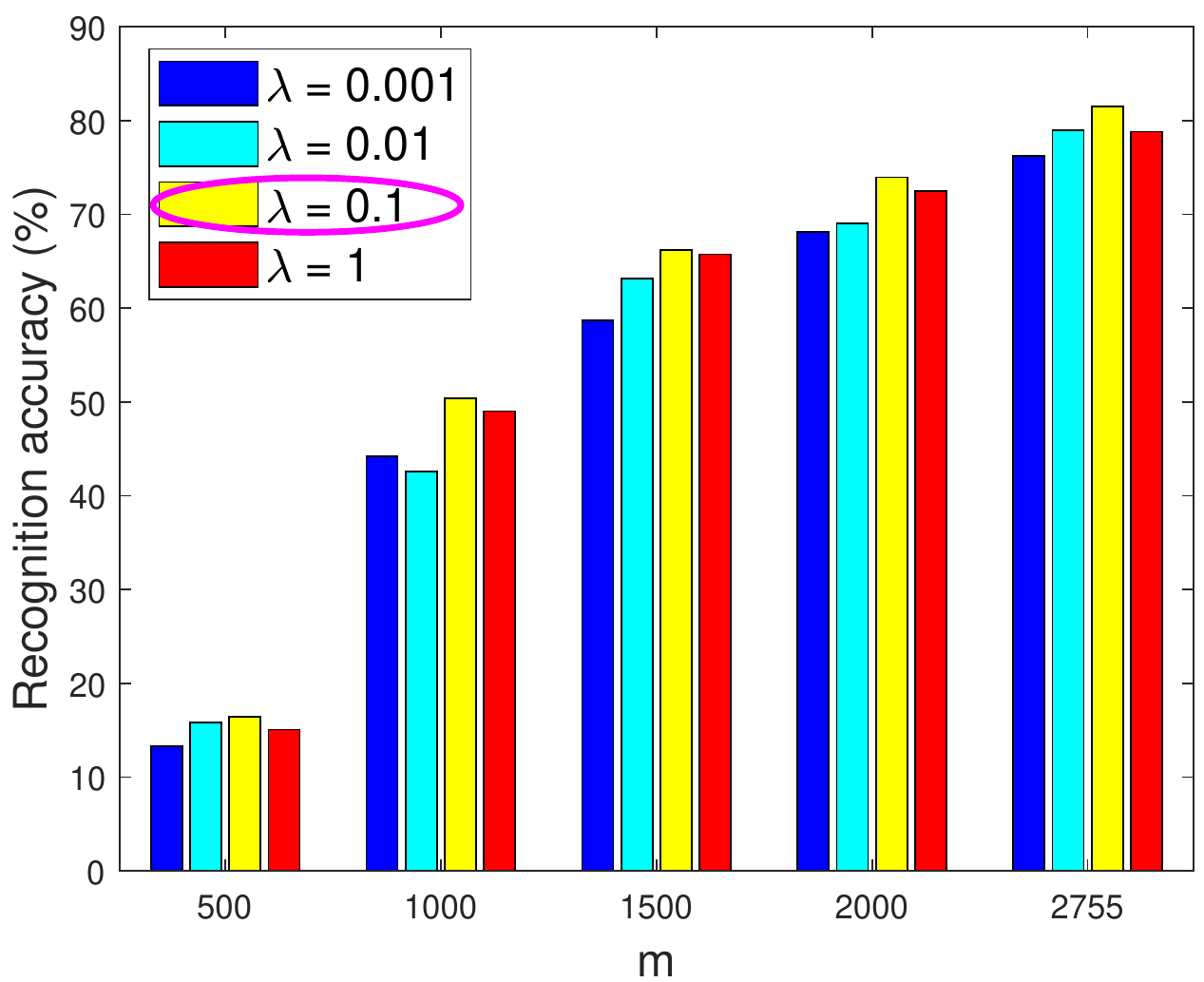}
    \end{center}
    \caption{The performance of the proposed method with respect to the hyperparameter $\lambda$ for the similarity loss.}
    \label{Fig:lambda-ablation}
\end{figure*}

\textbf{B. On choice of $\lambda$.}
We implemented a series of experiments over the printed artistic dataset in the character zero-shot settings with five different numbers $m$ of categories for training, i.e., $m\in \{500, 1000, 1500, 2000, 2755\}$ to determine the value of the tuning hyperparameter $\lambda$ for the similarity loss. Specifically, we considered four candidates of $\lambda \in \{0.001, 0.01, 0.1, 1\}$. The trends of recognition accuracy of the proposed method with different $\lambda$ are depicted in Figure \ref{Fig:lambda-ablation}. It can be observed from Figure \ref{Fig:lambda-ablation} that the proposed model achieves the best performance at the case of $\lambda = 0.1$ in all concerned zero-shot scenarios. Thus, in our experiments, we used 0.1 as the default value of $\lambda$.


\section{Conclusion}
\label{Conclusion}
This paper proposed a novel method for zero-shot Chinese character recognition by exploiting both stroke- and radical-level decomposition in the training stage and taking all the rectified stroke encodings with the nearest distance to enlarge the candidate set of characters in the inference stage. The effectiveness of the proposed method was demonstrated by numerous experiments over three benchmark datasets. The experimental results show that the proposed method outperforms the state-of-the-art methods in both character and radical zero-shot settings, and maintains the competitive performance in the traditional seen character settings. One future direction is to further improve the performance of the proposed method when applied to the zero-shot Chinese character recognition with complex backgrounds such as samples in the CTW dataset.

\section*{Acknowledgement}
The work of J. Zeng is partly supported by National Natural Science Foundation of China (No. 61977038) and the Thousand Talents Plan of Jiangxi Province (No. jxsq2019201124). 







\bibliographystyle{elsarticle-harv}
\bibliography{reference}
\end{document}